\documentclass[conference]{IEEEtran}
\IEEEoverridecommandlockouts
\usepackage{cite}
\usepackage{amsmath,amssymb,amsfonts}
\usepackage{algorithmicx}
\usepackage{algpseudocode}
\usepackage{algorithm}
\usepackage{algorithm}
\usepackage{graphicx}
\usepackage{epstopdf}
\usepackage{textcomp}
\usepackage{xcolor}
\usepackage{svg} 
\usepackage{booktabs}
\usepackage[table]{xcolor}
\usepackage{multirow}
\usepackage{threeparttable}
\usepackage{subcaption}
\usepackage{hyperref}
\usepackage[section]{placeins}
\def\BibTeX{{\rm B\kern-.05em{\sc i\kern-.025em b}\kern-.08em
    T\kern-.1667em\lower.7ex\hbox{E}\kern-.125emX}}
\begin{document}

\title{LLM-Based Agentic Exploration for Robot Navigation \& Manipulation with Skill Orchestration}

\author{\IEEEauthorblockN{Abu Hanif Muhammad Syarubany, Farhan Zaki Rahmani, Trio Widianto}
\IEEEauthorblockA{
\textit{Korea Advanced Institute of Science \& Technology (KAIST)}\\
Daejeon, South Korea\\
\{hanif.syarubany, farhanzaki19, trio.widi\}@kaist.ac.kr}}
% \and
% \and
% \and
% \and
% \and

\maketitle

\begin{abstract}
This paper presents an end-to-end LLM-based agentic exploration system for an indoor shopping task, evaluated in both Gazebo simulation and a corresponding real-world corridor layout. The robot incrementally builds a lightweight semantic map by detecting signboards at junctions and storing direction-to-POI relations together with estimated junction poses, while AprilTags provide repeatable anchors for approach and alignment. Given a natural-language shopping request, an LLM produces a constrained discrete action at each junction (direction and whether to enter a store), and a ROS finite-state main controller executes the decision by gating modular motion primitives, including local-costmap-based obstacle avoidance, AprilTag approaching, store entry, and grasping. Qualitative results show that the integrated stack can perform end-to-end task execution from user instruction to multi-store navigation and object retrieval, while remaining modular and debuggable through its text-based map and logged decision history.
\end{abstract}


\section{Introduction}
Recent progress in visual perception, semantic mapping, and large language models (LLMs) is enabling mobile robots to follow high-level instructions rather than fixed waypoints. The key challenge is integration: a robot must perceive discrete cues, maintain a semantic environment, and translate language-level goals into safe, executable motions in real-time.

In this project, we study an agentic exploration task in a corridor-based shopping environment with multiple store categories and a pickup point, evaluated in both Gazebo~\cite{koenig2004gazebo} and a corresponding real-world setup. Information about store locations is not given as ground-truth coordinates; instead, it is distributed across junction signboards containing directional arrows, store icons, and AprilTags~\cite{olson2011apriltag}. The robot must explore, build a semantic map from these cues, and decide which direction to take at each junction to satisfy a user-provided shopping order.

We implement an end-to-end ROS stack~\cite{quigley2009ros} that couples an LLM decision layer with modular low-level controllers. ORB-SLAM3~\cite{campos2021orbslam3} provides global pose, AprilTags~\cite{olson2011apriltag} act as stable anchors for approach behaviors, and perception nodes write a lightweight JSON semantic map of junction-to-POI relations. Given the mission, current junction context, and a history log, an LLM outputs a constrained discrete command \texttt{<direction>|||<store\_action>}, which a finite-state \textit{MainController} executes by gating motion primitives (wall avoidance, tag approach, store entry, and grasping).

\section{Related Work}
Metric localization and mapping are core requirements for indoor mobile robots. Feature-based visual SLAM systems such as ORB-SLAM provide accurate monocular pose tracking with keyframe mapping and loop closure \cite{murartal2015orbslam}. ORB-SLAM2 extends this pipeline to stereo and RGB-D sensing while maintaining real-time performance \cite{murartal2017orbslam2}. ORB-SLAM3 further improves robustness by supporting visual inertial sensing and multi-map operation, which is useful in re-localization after tracking loss \cite{campos2021orbslam3}. In our stack, we rely on SLAM to provide a continuous global pose estimate that downstream modules can use for logging motion primitives.

Fiducial markers are widely used to complement SLAM with strong, repeatable geometric anchors. AprilTag provides a robust, open fiducial system that enables accurate pose estimation from a single camera view \cite{olson2011apriltag}, and AprilTag~2 improves detector efficiency and robustness in practical settings \cite{wang2016apriltag2}. In our scenario, AprilTags embedded in signboards serve as reliable junction/entrance references that help stabilize the semantic map and support precise approach behaviors.

Beyond purely geometric maps, semantic representations associate places with symbolic meaning (e.g., store categories) so that tasks can be solved in terms of “what” and “where” rather than only metric coordinates. Surveys on semantic mapping in mobile robotics summarize common formulations and the trade-offs between metric, topological, and hybrid representations \cite{kostavelis2015semantic, garg2021semantics}. Our approach follows this line by maintaining a lightweight, text-based semantic map of junctions, directions, and store categories, which is convenient for both debugging and for prompting an LLM with structured context.

Recent work also explores large language models as high-level planners or policies for embodied agents. One direction is to use language models to infer feasible action sequences or subgoals from structured state descriptions \cite{huang2022zeroshot}. Other approaches explicitly ground language to robot affordances (e.g., mapping instructions into executable skills under constraints) \cite{ahn2022saycan}, or generate code-like policies that call lower-level controllers \cite{liang2023code}. In contrast to end-to-end learned navigation, our project uses an LLM only as a discrete decision layer over a junction-store semantic graph; low-level motion is executed by separate ROS controllers (wall following, tag approaching, store entry, recovery), enabling modularity and safer execution.

\begin{figure*}[htbp]
    \centering
    \includegraphics[width=\textwidth]{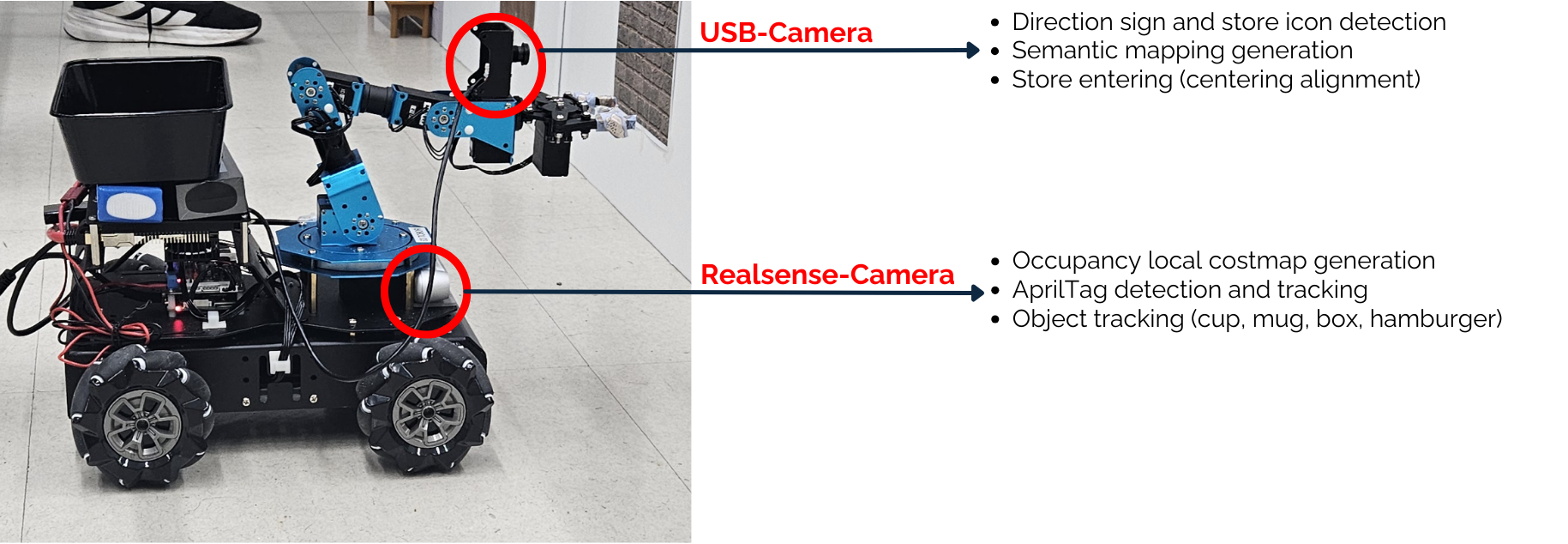}
    \caption{\textbf{Camera setup and responsibilities used in the proposed system.} The USB camera is used for signboard understanding and store-entry centering, whereas the RealSense depth camera supports local costmap construction, AprilTag detection, and object tracking.}

    \label{fig:robot_camera_functions}
\end{figure*}

\begin{figure*}[htbp]
    \centering
    \includegraphics[width=0.65\textwidth]{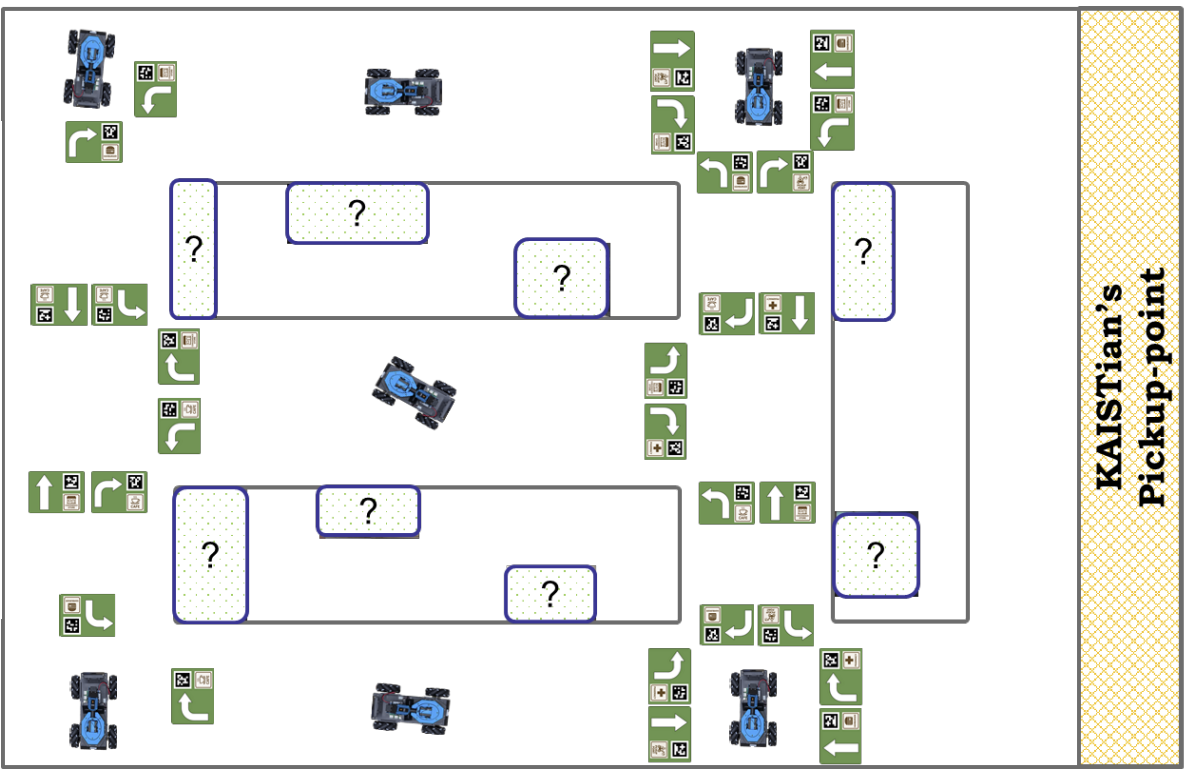}
    \caption{\textbf{Map layout of the corridor-based shopping environment.} Junction signboards (arrows, store icons, and AprilTags) support semantic mapping and navigation, and the shaded region on the right marks the pickup point.}

    \label{fig:map_layout}
\end{figure*}

\section{Methodology}

\begin{figure}[H]
    \centering
    \begin{minipage}{0.5\columnwidth}
        \centering
        {\includegraphics[width=\linewidth]{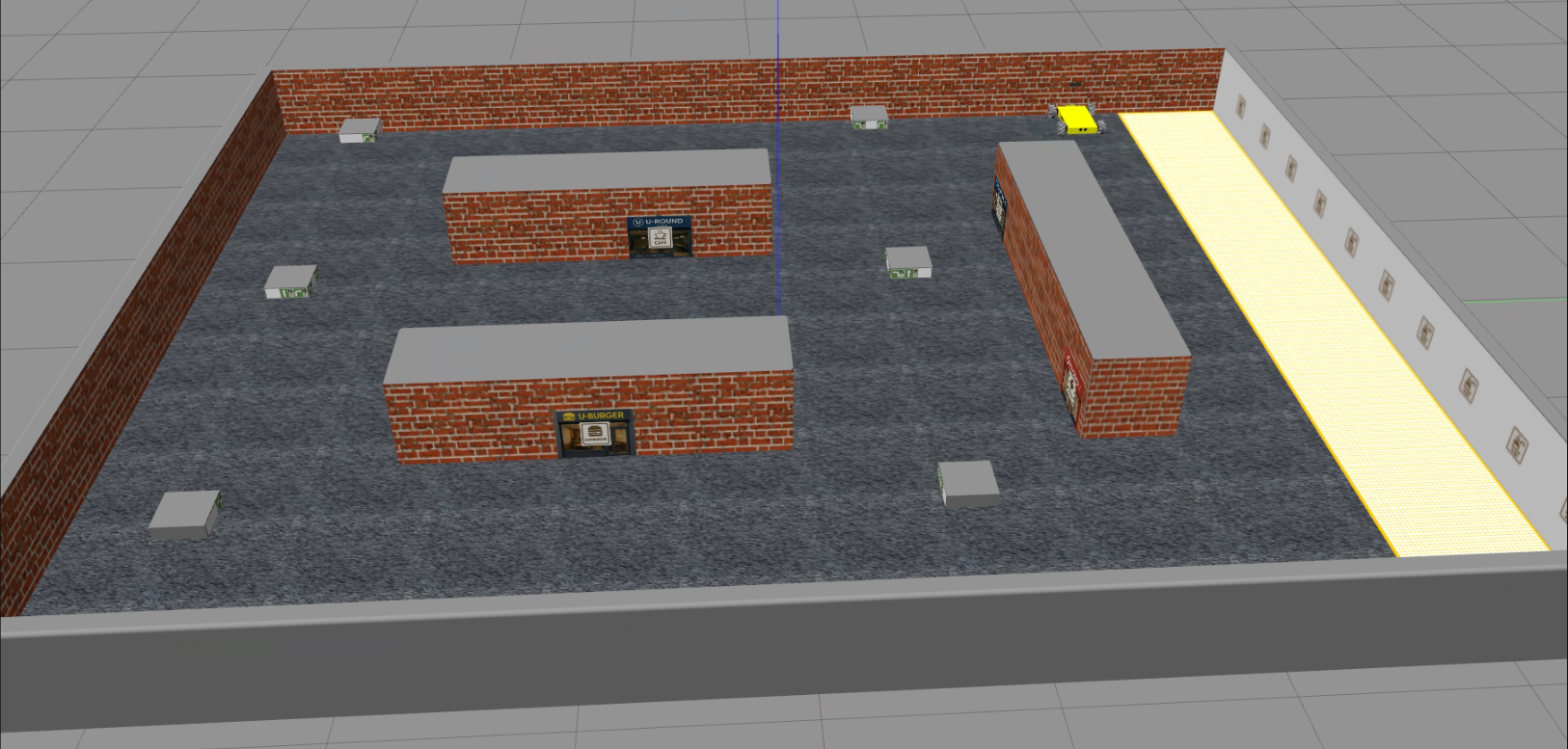}}
        \caption*{(a) Simulated map}
    \end{minipage}
    \begin{minipage}{0.42\columnwidth}
        \centering
        {\includegraphics[width=\linewidth]{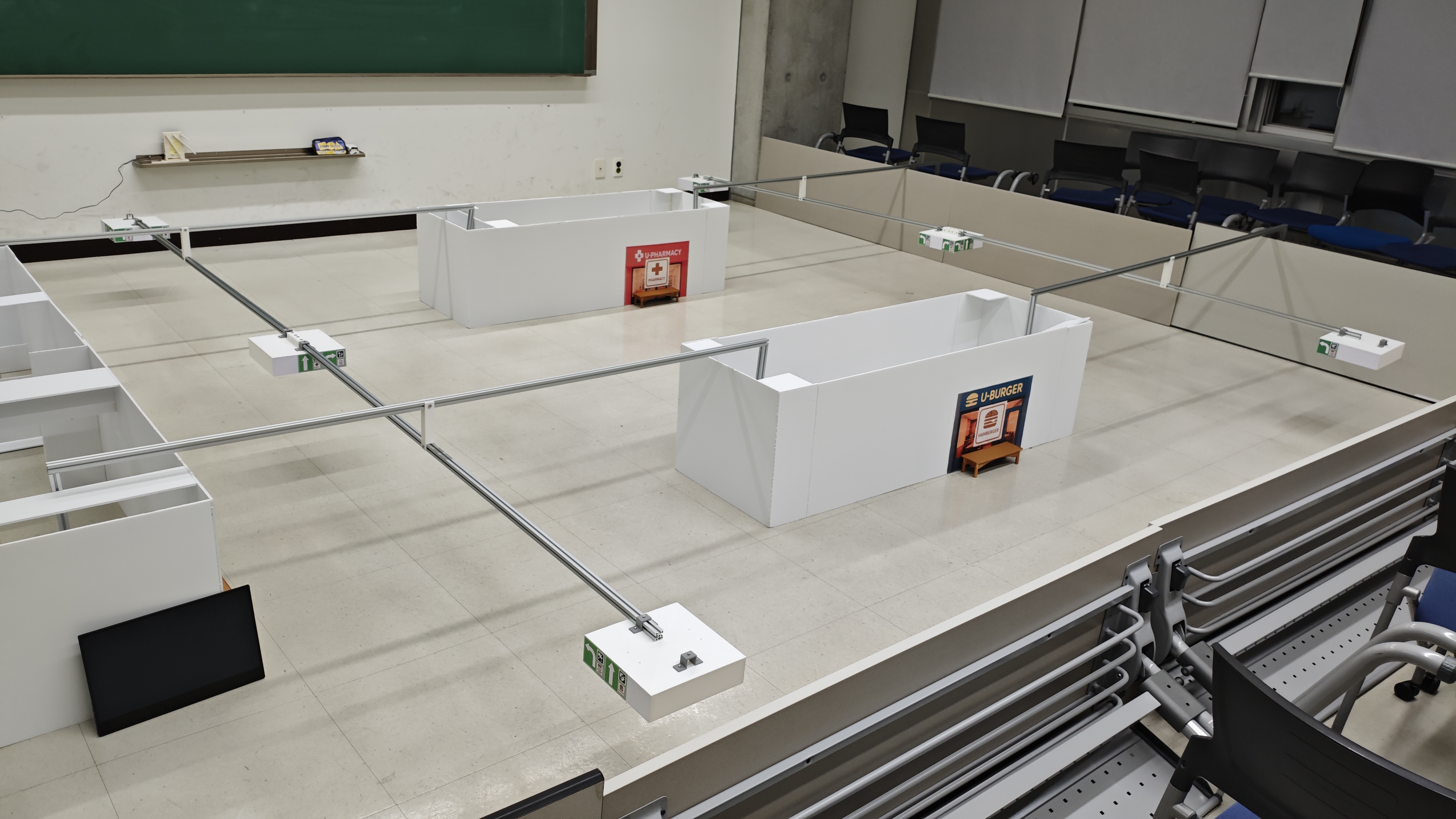}}
        \caption*{(b) Real map}
    \end{minipage}
    \caption{\textbf{Environment layouts used in this work.} (a) Gazebo simulation and (b) the corresponding real-world setup. Both share the same corridor structure and junction signboards for evaluation.}
    \label{fig:env_layout}
\end{figure}

We developed our agentic exploration pipeline in a corridor-based shopping environment with consistent junction structure, using both a Gazebo simulation~\cite{koenig2004gazebo} and a corresponding real-world layout (Fig.~\ref{fig:map_layout} \& Fig.~\ref{fig:env_layout}). The system is designed to build a semantic map from signboard cues and execute language-conditioned tasks through an LLM decision layer coupled with ROS motion primitives. Our sensing setup uses a two-camera configuration (Fig.~\ref{fig:robot_camera_functions}), where an RGB USB camera supports signboard/store-icon perception and entry centering, while a RealSense RGB-D camera provides depth for local costmap construction, AprilTag perception~\cite{olson2011apriltag}, and grasp-target tracking.

\begin{table}[t]
\centering
\caption{Distribution of training and validation images per class for the YOLO dataset.}
\label{tab:yolo_class_distribution}
\begin{tabular}{lccc}
\toprule
\textbf{Class} & \textbf{Train} & \textbf{Validation} & \textbf{Total} \\
\midrule
Negative            & 1335 (4.6\%)  & 340 (4.6\%)  & 1675 (4.6\%)  \\
Cup                 & 3258 (11.3\%) & 838 (11.2\%) & 4096 (11.3\%) \\
Mug                 & 3287 (11.4\%) & 853 (11.4\%) & 4140 (11.4\%) \\
Hamburger           & 4851 (16.8\%) & 1258 (16.9\%)& 6109 (16.8\%) \\
Box                 & 3283 (11.4\%) & 850 (11.4\%) & 4133 (11.4\%) \\
Direction-Left                & 1081 (3.7\%)  & 278 (3.7\%)  & 1359 (3.7\%)  \\
Direction-Right               & 1105 (3.8\%)  & 286 (3.8\%)  & 1391 (3.8\%)  \\
Direction-Straight            & 1142 (4.0\%)  & 292 (3.9\%)  & 1434 (3.9\%)  \\
Pickup point           & 1581 (5.5\%)  & 409 (5.5\%)  & 1990 (5.5\%)  \\
Cafe                & 1644 (5.7\%)  & 428 (5.7\%)  & 2072 (5.7\%)  \\
Convenience store & 1574 (5.5\%)  & 406 (5.4\%)  & 1980 (5.5\%)  \\
Hamburger store   & 1575 (5.5\%)  & 408 (5.5\%)  & 1983 (5.5\%)  \\
Pharmacy         & 1602 (5.5\%)  & 413 (5.5\%)  & 2015 (5.5\%)  \\
Apriltag                  & 1557 (5.4\%)  & 395 (5.3\%)  & 1952 (5.4\%)  \\
\midrule
\textbf{Total}      & 28875 (100\%) & 7454 (100\%) & 36329 (100\%) \\
\bottomrule
\end{tabular}
\end{table}

\begin{figure*}[htbp]
    \centering
    \includegraphics[width=\textwidth]{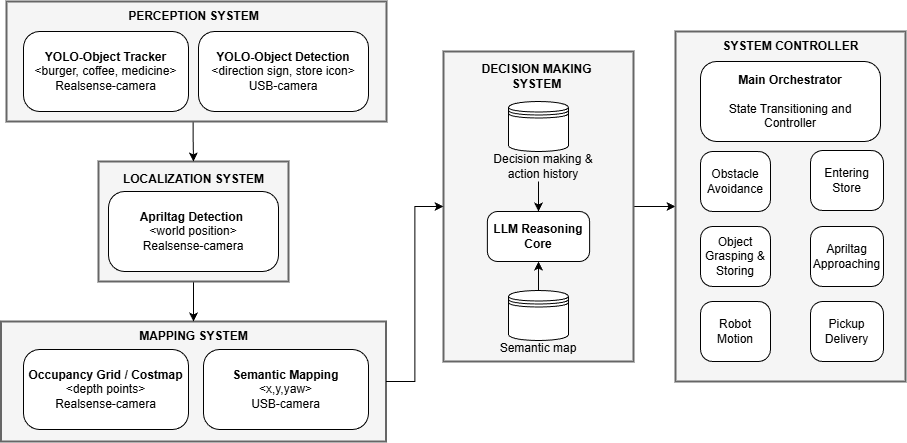}
    \caption{\textbf{System-level pipeline of our agentic exploration stack, including perception, localization, mapping, and an LLM reasoning core.} The controller layer uses the semantic map and decision history to trigger low-level skills through a finite-state orchestrator.}
    \label{fig:system_pipeline}
\end{figure*}

\begin{figure}[t]
    \centering
    \begin{minipage}{0.5\columnwidth}
        \centering
        {\includegraphics[width=\linewidth]{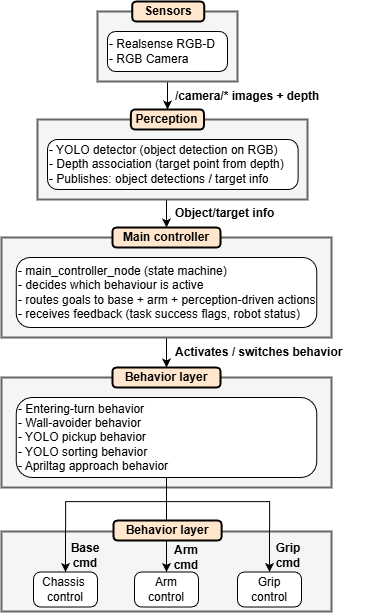}}
        \caption*{(a) Architecture flow}
    \end{minipage}
    \begin{minipage}{0.404\columnwidth}
        \centering
        {\includegraphics[width=\linewidth]{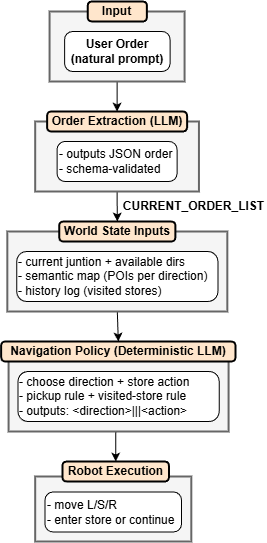}}
        \caption*{(b) LLM agent flow}
    \end{minipage}
    \caption{\textbf{Software pipeline of the proposed system.} (a) ROS architecture flow from sensors and perception. (b) LLM agent flow, where a natural-language order to produce a constrained action \texttt{<direction>|||<store\_action>} for robot execution.}
    \label{fig:software_pipeline}
\end{figure}

\begin{algorithm}[t]
\caption{Gated AprilTag approaching with final forward advance.}
\label{alg:apriltag_approach}
\footnotesize
\begin{algorithmic}[1]
\State \textbf{Input:} tag detections $(x,z,\psi)$, gate $g$, targets $(x^{\ast}, d^{\ast})$, tolerances $(\tau_x,\tau_z,\tau_\psi)$
\State \textbf{Params:} gains $(k_x,k_z,k_\psi)$, limits $(v_{x,\max},v_{y,\max},\omega_{\max})$, timeout $T$, advance $(d_{\text{adv}},v_{\text{adv}})$
\State \textit{active}$\leftarrow$False; \textit{adv\_done}$\leftarrow$False; $t_{\text{det}}\leftarrow -\infty$
\If{$g=1$} \State \textit{active}$\leftarrow$True; \textit{adv\_done}$\leftarrow$False \EndIf
\If{$g=0$} \State \textit{active}$\leftarrow$False; publish stop; publish success$=0$ \EndIf
\While{ROS running}
\If{\textbf{not} \textit{active}} \State \textbf{continue} \EndIf
\If{new tag} \State update $(x,z,\psi)$; $t_{\text{det}}\leftarrow \text{now}$ \EndIf
\If{$\text{now}-t_{\text{det}} > T$} \State publish stop; publish success$=0$; \textbf{continue} \EndIf
\State $e_x\!\leftarrow\!x-x^{*}$,\; $e_z\!\leftarrow\!z-d^{*}$,\; $e_\psi\!\leftarrow\!\psi$
\If{$|e_x|>\tau_x$ \textbf{or} $|e_z|>\tau_z$}
\State $v_y\leftarrow \mathrm{sat}(-k_x e_x,\pm v_{y,\max})$, $v_x\leftarrow \mathrm{sat}(k_z e_z,\pm v_{x,\max})$
\State $\omega\leftarrow \mathrm{sat}(k_\psi e_\psi,\pm \omega_{\max})$; publish $(v_x,v_y,\omega)$; success$=0$
\ElsIf{$|e_\psi|>\tau_\psi$}
\State $\omega\leftarrow \mathrm{sat}(k_\psi e_\psi,\pm \omega_{\max})$; publish $(0,0,\omega)$; success$=0$
\ElsIf{\textbf{not} \textit{adv\_done}}
\State drive forward $v_x=\min(|v_{\text{adv}}|,v_{x,\max})$ for $d_{\text{adv}}/v_x$ seconds; then stop
\State \textit{adv\_done}$\leftarrow$True; publish success$=1$
\EndIf
\EndWhile
\end{algorithmic}
\end{algorithm}

\begin{algorithm}[t]
\caption{Local-costmap wall avoidance with gated FSM.}
\label{alg:wall_avoider}
\footnotesize
\begin{algorithmic}[1]
\State \textbf{Input:} local occupancy grid $M$, gate $g\in\{0,1\}$, parameters $(d_{\text{safe}}, v_f, \omega, \theta_{90}, \theta_{180})$
\State \textbf{State:} $s\in\{\textsc{Forward},\textsc{Turn90},\textsc{Turn180}\}$, last tag time $t$
\State $s\leftarrow \textsc{Forward}$; publish \texttt{/wall\_avoider\_state}$\leftarrow s$
\While{ROS running}
\If{$g=0$} \State publish stop once; $s\leftarrow\textsc{Forward}$; \textbf{continue} \EndIf
\State $d \leftarrow \Call{DistAhead}{M}$ \Comment{min obstacle distance in a forward corridor slice}
\If{$s=\textsc{Forward}$}
    \If{$d\neq \emptyset$ \textbf{and} $d\le d_{\text{safe}}$} \State $s\leftarrow\textsc{Turn90}$; $t\leftarrow\text{now}$ \Else \State publish $(v_f,0)$ \EndIf
\ElsIf{$s=\textsc{Turn90}$}
    \If{$\text{now}-t < \theta_{90}/|\omega|$} \State publish $(0,\pm\omega)$
    \Else
        \State publish stop; $d'\leftarrow \Call{DistAhead}{M}$
        \If{$d'\neq \emptyset$} \State $s\leftarrow\textsc{Turn180}$; $t\leftarrow\text{now}$ \Else \State $s\leftarrow\textsc{Forward}$ \EndIf
    \EndIf
\ElsIf{$s=\textsc{Turn180}$}
    \If{$\text{now}-t < \theta_{180}/|\omega|$} \State publish $(0,\pm\omega)$
    \Else \State $s\leftarrow\textsc{Forward}$; publish $(v_f,0)$ \EndIf
\EndIf
\State publish \texttt{/wall\_avoider\_state}$\leftarrow s$
\EndWhile
\end{algorithmic}
\end{algorithm}

\begin{algorithm}[t]
\caption{MainController FSM (latched gating of motions).}
\label{alg:main_controller}
\footnotesize
\begin{algorithmic}[1]
\State Disable all gates; $S\gets$ \textsc{WallAvoid}
\While{ROS running}
\If{$g_m=0$} \State Disable all; publish stop; $S\gets$ \textsc{WallAvoid}; \textbf{continue} \EndIf

\If{$S=\textsc{WallAvoid}$}
    \State Enable wall-avoid gate
    \If{tag recent $\land$ not turning $\land$ not in cooldown}
        \State $S\gets$ \textsc{TagApproach}
    \EndIf
\ElsIf{$S=\textsc{TagApproach}$}
    \State Enable AprilTag-approach gate
    \If{approach success} \State Start mapping timer; $S\gets$ \textsc{Map} \EndIf
    \If{tag lost $\land$ last distance $> d_{\text{far}}$} \State $S\gets$ \textsc{WallAvoid} \EndIf
\ElsIf{$S=\textsc{Map}$}
    \State Enable semantic-mapping gate
    \If{elapsed $\ge T_{\text{map}}$} \State Stop mapping; $S\gets$ \textsc{WaitAction} \EndIf
\ElsIf{$S=\textsc{WaitAction}$}
    \If{received action $a=(dir,target)$} \State Trigger pre-enter$(dir)$; $S\gets$ \textsc{PreEnter} \EndIf
\ElsIf{$S=\textsc{PreEnter}$}
    \If{$r_p\neq\texttt{success}$} \State $S\gets$ \textsc{WallAvoid} \EndIf
    \If{$r_p=\texttt{success}$}
        \If{$target=\texttt{pickup}$} \State Enable pickup gate; $S\gets$ \textsc{Pickup} \EndIf
        \If{$target=\texttt{continue}$} \State $S\gets$ \textsc{WallAvoid} \EndIf
        \State Enable entering-store$(target)$; $S\gets$ \textsc{EnterStore}
    \EndIf
\ElsIf{$S=\textsc{EnterStore}$}
    \If{$r_e=\texttt{success}$} \State Enable grasp gate; $S\gets$ \textsc{Grasp} \EndIf
    \If{$r_e\neq\texttt{success}$} \State $S\gets$ \textsc{WallAvoid} \EndIf
\ElsIf{$S=\textsc{Grasp}$}
    \If{$s_g=1$} \State Post-grasp turn (opposite $dir_{\text{prev}}$) then timed backup; $S\gets$ \textsc{WallAvoid} \EndIf
\ElsIf{$S=\textsc{Pickup}$}
    \State Keep pickup gate ON forever; ignore new actions
\EndIf
\EndWhile
\end{algorithmic}
\end{algorithm}

\subsection{Object Detection Development}
To support both signboard understanding and grasp-target perception, we constructed a custom YOLO~\cite{yolo11_ultralytics} dataset from simulation and real-environment runs by logging camera frames across diverse viewpoints, lighting conditions, and levels of clutter, then filtering for representative positives and hard negatives. To reduce manual annotation effort while retaining strong supervision, we adopt a weak distillation pipeline: we distill semantic and localization knowledge from a large vision-language model (Qwen~2.5-VL~\cite{bai2025qwen25vltechnicalreport}) into a lightweight YOLO detector (YOLO11-small~\cite{yolo11_ultralytics}) by generating pseudo-labels (class tags and bounding boxes) and then refining them through simple sanity checks (e.g., size/aspect constraints and duplicate removal). The resulting dataset spans grasp targets (cup, mug, hamburger, box), directional arrows (left/right/straight), semantic cues (store icons and pickup point), and AprilTags  (Table~\ref{tab:yolo_class_distribution}), where the class counts are moderately imbalanced (e.g., hamburger dominates among grasp objects), reflecting the task distribution in our environment. Importantly, we include a substantial number of negative samples to expose the detector to visually similar but non-actionable scenes, which reduces false positives and improves stability when triggering downstream behaviors (e.g., store entry alignment and grasping). We train and deploy YOLO11-small as the real-time detector used throughout our system.

\subsection{System Architecture Overview}

Our system is an end-to-end ROS stack that connects perception, semantic mapping, LLM-based decision-making, and low-level execution in a simulated shopping-mall environment. As summarized in Fig.~\ref{fig:system_pipeline}, raw sensor streams are processed by perception and localization modules to produce (i) a local metric representation for safe motion and (ii) a symbolic semantic map that captures junction-to-store relations for high-level reasoning.

The front-end consists of two perception pipelines: object detection/tracking for grasp targets and signboard understanding for navigation cues, together with AprilTag-based localization to provide reliable junction/entrance anchors (Fig.~\ref{fig:system_pipeline}). In parallel, a mapping layer maintains an occupancy-based local costmap for obstacle avoidance and a lightweight semantic map (stored as JSON) that records each discovered junction with direction-to-POI pairs and an estimated 2D pose.

The control back-end follows the layered architecture in Fig.~\ref{fig:software_pipeline}(a). A \textit{MainController} finite-state machine orchestrates modular behaviors (wall avoidance, AprilTag approaching, store entry, grasping, and recovery) via latched gating topics, allowing only one skill to command \texttt{/cmd\_vel} at a time. This design supports robust switching between behaviors without requiring an end-to-end learned controller.

Finally, the LLM agent flow is shown in Fig.~\ref{fig:software_pipeline}(b). A natural-language user order is first converted into a schema-validated structured order list, then combined with the current junction context, the semantic map, and an action-history log to produce a constrained discrete command of the form \texttt{<direction>|||<store\_action>}. The MainController consumes this command to trigger the appropriate low-level behaviors for navigation and manipulation while keeping safety and motion execution local and deterministic.

\subsection{Perception and Semantic Mapping}

\subsubsection{Signboard and AprilTag perception}
The Gazebo world contains signboards at major junctions. Each signboard encodes a junction identifier via its pose, one or more directional arrows, store–category icons, and an AprilTag that serves as a robust fiducial anchor. Dedicated perception nodes (provided by the group) subscribe to RGB–D camera topics and to the AprilTag detection topic, and convert each signboard observation into a symbolic junction record. A record contains the robot pose when the signboard was observed (in the odometry frame), a junction name, and a list of “POI pairs". These records are published on a semantic–mapping topic and appended to a persistent JSON file.
\subsubsection{Semantic map representation}
The semantic map is stored as a list of junction entries. Each entry includes a unique identifier (e.g., \texttt{"junction\_3"}), a list of direction–POI strings, and the estimated 2D pose \texttt{(x, y, yaw)} of the junction at the time of discovery. This representation deliberately separates symbolic knowledge (“left leads to a pharmacy”) from metric knowledge (approximate junction pose). It is simple enough to inspect or edit by hand, yet rich enough for the LLM navigation policy to plan over junctions, store types, and the pickup point without dealing with raw geometry or continuous trajectories.
\subsubsection{Historical trajectory logging}
To support agentic behavior and post–hoc analysis, a historical–mapping node subscribes to both semantic events and high–level action decisions. Whenever the robot stops at a junction or executes a turn command, the node writes a textual log entry into a history JSON file. Over time this file describes which junctions have been visited, which directions were taken, and which store instances were entered. The navigation LLM uses this log to avoid revisiting the same store instance unnecessarily, to reason about untried branches, and to reconstruct how close the mission is to completion.

\subsection{LLM-Based Navigation and Task Parsing}
\begin{algorithm}[t]
\caption{LLM navigation rule.}
\label{alg:llm_nav_policy}
\footnotesize
\begin{algorithmic}[1]
\Require Semantic map $\mathcal{M}$, current junction $J$, remaining order $R$, history log $H$
\Ensure Output only \texttt{<dir>|||<act>} with $dir\in\{1,2,3\}$ (L/S/R) and $act\in\{1,2,3,4,5\}$ (store/continue)

\State From $H$, extract visited store-instances $(\texttt{junction},\texttt{dir},\texttt{poi})$ and tried branches $(\texttt{junction},\texttt{dir})$
\If{$R=0$} \Comment{Shopping finished}
    \State Go toward \texttt{pickup} (if at $J$, take its direction); output \texttt{<dir>|||5}
\Else \Comment{Shopping ongoing}
    \State Select best unvisited store-instance in $\mathcal{M}$ by matching store type to remaining items in $R$
    \If{that instance is at $J$} \State Enter now: output \texttt{<dir>|||<store\_id>} \EndIf
    \State Otherwise move: head toward the target (or pick an untried branch if no useful store is known)
    \State Avoid \texttt{pickup} directions while $R>0$ if an alternative exists; output \texttt{<dir>|||5}
\EndIf
\end{algorithmic}
\end{algorithm}

\begin{figure*}[htbp]
    \centering
    \includegraphics[width=\textwidth]{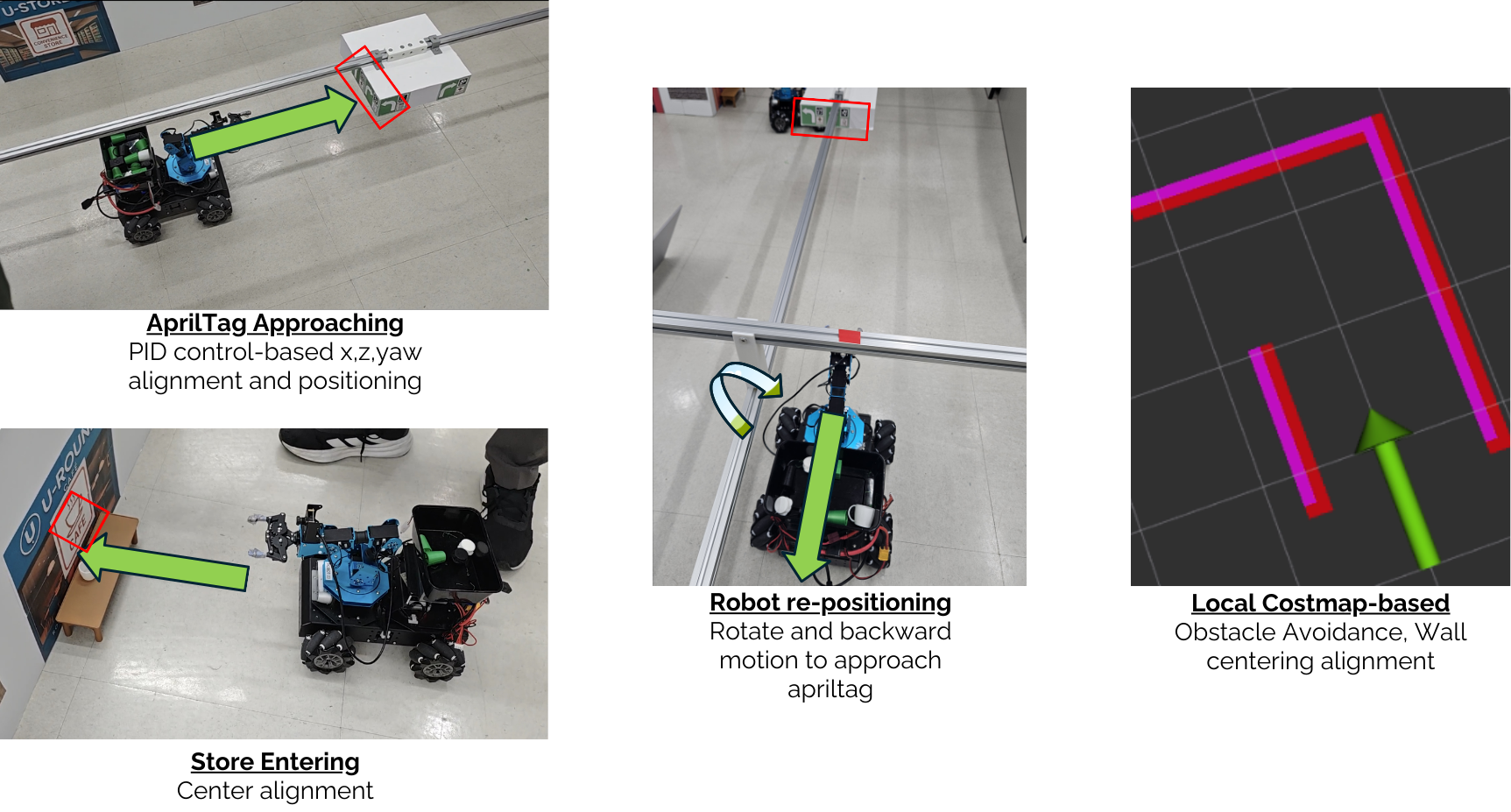}
    \caption{\textbf{Representative motion strategies used during exploration and semantic-map building}, including AprilTag-based approaching, base re-positioning, local-costmap obstacle avoidance, and store-entry alignment. These primitives are triggered by the main controller to reliably reach signboards and entrances while maintaining safe navigation in corridors.}
    \label{fig:motion_strategies}
\end{figure*}
\subsubsection{Order parsing from natural language}
At the start of a run, the user specifies a shopping task in natural language on a dedicated ROS topic (for example, “bring two hamburgers and one emergency medicine”). A Python node subscribes to this topic and forwards the raw text to an LLM prompt that is constrained by a JSON schema. The model must output a dictionary with four non–negative integers: the requested counts of hamburgers, iced coffee, hot coffee, and medicine. The node post–processes the result, enforces basic sanity checks, and saves the normalized order list to a JSON file. This file is treated as the ground–truth shopping list and is read both by the navigation policy and by the object–grasping module.
\subsubsection{Navigation policy over the semantic map}
The core of the decision layer is the \texttt{DecisionMakers} node. Whenever the robot finishes semantic mapping at a junction, this node gathers: (i) the current junction entry from the semantic map, (ii) the full list of discovered junctions and their POI pairs, (iii) the current order list, and (iv) the text history of previous decisions from the trajectory log. It then constructs a structured prompt and calls the navigation LLM.

The system prompt (Algorithm~\ref{alg:llm_nav_policy}) defines a deterministic navigation policy. The LLM must output a discrete command of the form
\texttt{<DIRECTION>|||<STORE\_ACTION>}
where the first code chooses among left / straight / right and the second code chooses whether to enter a specific store type now or to continue searching. The prompt also specifies how to treat different store capabilities, how to treat individual store instances (identified by junction name, direction, and POI name), and when to prioritize the pickup point once all requested items have been collected. The node parses the two integer codes, validates that the chosen direction exists in the current junction, and finally publishes a textual command. Importantly, the LLM never produces velocity commands; it only decides which semantic branch to take and whether to enter a store at that branch.

\subsection{Low-Level Motion Execution and Recovery}
\subsubsection{Motion primitives and local costmap controller}
Low-level motion is handled by classical ROS controllers that operate on \texttt{/cmd\_vel} and use a local costmap for safety. A local–costmap node fuses depth or ultrasonic readings into a 2D occupancy grid around the robot. Cells occupied by nearby obstacles receive high cost, and free space in front of the robot appears as low cost. The wall–avoidance controller (Algorithm~\ref{alg:wall_avoider}) consumes this local costmap together with the desired direction (left / straight / right) and generates smooth velocity commands that move the robot along that branch while keeping a safe distance to obstacles.

Additional motion primitives include an AprilTag–approach (Algorithm~\ref{alg:apriltag_approach}) controller that aligns and stops the robot in front of a signboard or store entrance, a “pre–entering” controller that centers the robot with respect to a selected doorway direction, and an entering–store controller that drives the robot into the store and stops in a stable pose. These primitives are implemented as small state machines that send velocity commands at a fixed rate, using odometry and range measurements rather than any LLM feedback.
\subsubsection{Main controller and recovery state}
The \texttt{MainController} node orchestrates all motion primitives using a finite–state machine as stated in Algorithm~\ref{alg:main_controller}. When the main gate topic is activated, the controller starts in a \textsc{WALL\_AVOID} state: it enables the wall–avoidance node (and therefore the local costmap) so that the robot can safely explore corridors until an AprilTag is detected. Upon seeing a nearby tag, it switches to an \textsc{APRILTAG\_APPROACH} state (Algorithm~\ref{alg:apriltag_approach}), then to \textsc{SEMANTIC\_MAPPING} to let the perception node update the semantic map. After mapping, the controller enters \textsc{WAIT\_NEXT\_ACTION} and blocks until a new high–level command from the LLM arrives on \texttt{/next\_controller\_action}. Depending on this command, as illustrated in the Figure~\ref{fig:motion_strategies}, it triggers the pre–enter, entering–store, multi–object grasping, post–grasp turning, or pickup behaviors, each with its own success topic.

To improve robustness, the controller monitors simple progress indicators such as success flags from each primitive and timeouts. If a primitive reports failure or the robot fails to make progress, the controller falls back to wall–avoidance with the local costmap or activates a special pickup state that hands control to a dedicated pickup controller. In every case, the recovery logic remains local and reactive; once the robot reaches a new junction and semantic mapping is complete, control is passed back to the LLM navigation policy via the next–action topic.

\section{Experiments and Results}
\begin{figure}[t]
    \centering
    \fbox{\includegraphics[width=\columnwidth]{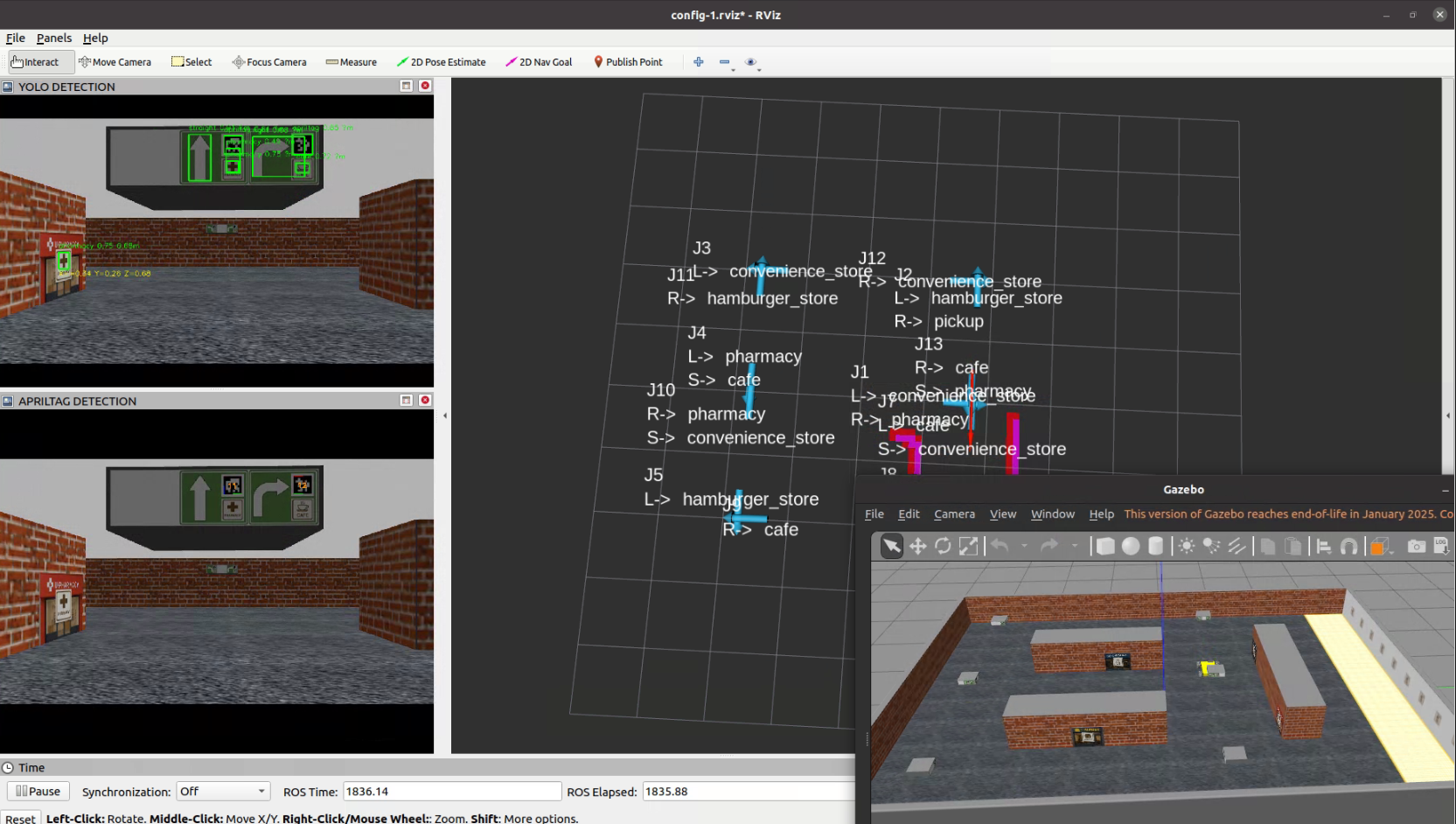}}
    \caption{\textbf{System overview during an exploration run.} Showing RViz for semantic-map and trajectory visualization alongside the Gazebo simulation.}
    \label{fig:rviz_overview}
\end{figure}

\subsection{End-to-End Task Execution}
We evaluated the full pipeline in both a Gazebo simulation and a corresponding real-world environment containing multiple stores and a pickup point (Fig.~\ref{fig:rviz_overview}). At the beginning of each episode, the robot starts near the pickup zone with an empty semantic map. During an initial exploration phase, the wall-avoidance controller uses the local costmap to drive the robot through the corridors while keeping a safe distance to obstacles, and the perception nodes detect signboards at junctions and record both the junction pose and the symbolic relations between directions and store categories in a shared JSON map.

Once a user goal is provided, the system performs a complete end-to-end task from natural-language instruction to object pickup. The LLM-based navigation policy reads the current semantic map and the history of visited locations, and outputs the next high-level action at each junction. The resulting command is forwarded to the main controller, which triggers the appropriate motion primitives: turning into the chosen branch, following the corridor using the local costmap, approaching the next AprilTag, and, when required, running the pre-entering and entering-store routines. When the robot reaches the vicinity of a target shelf, a separate grasping controller is triggered to align the end-effector and perform object pickup.

Qualitatively, the integrated system was able to complete multi-store missions in both simulation and the real setup: after a short exploration phase, the robot could visit the required store categories in a reasonable order and then return near the starting area. The clear separation between high-level semantic decisions (LLM) and low-level motion execution (local-costmap-based navigation and grasping node) allowed us to change or refine the policy logic without modifying the underlying controllers.

\subsection{Semantic Mapping}
\begin{figure}[t]
    \centering
    \includegraphics[width=\columnwidth]{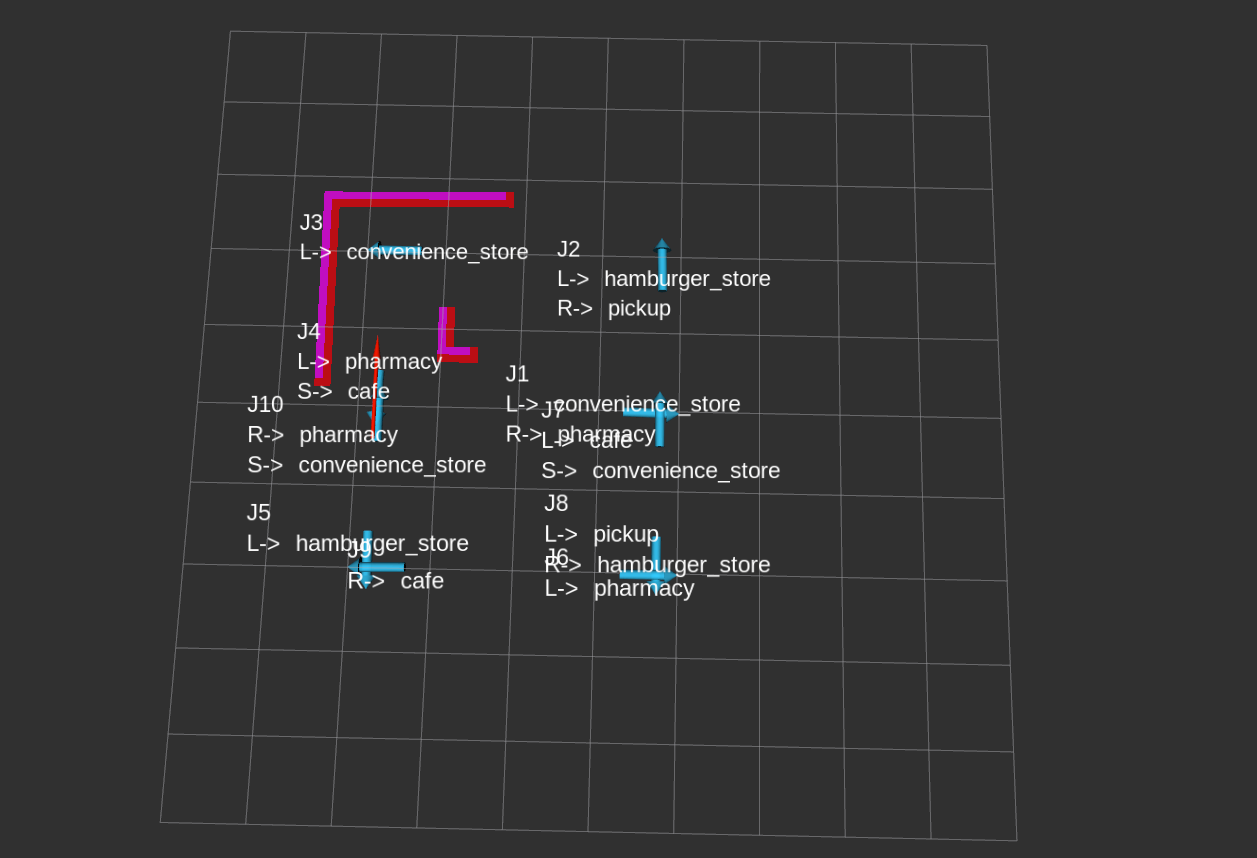}
    \caption{\textbf{Semantic map built online during exploration.} Each junction stores directional relations to store categories (Left/Straight/Right).}
    \label{fig:semantic_map}
\end{figure}

The semantic mapping module aggregates information from signboards detected at junctions into a compact JSON representation. Each entry stores the robot pose at the time of observation and a list of directional relations. Because this representation is independent of any particular controller, it can be reused both by the LLM navigation policy and by debugging tools.

In practice, the map grew incrementally as the robot explored new corridors under wall-avoidance control (Fig.~\ref{fig:semantic_map}). Once a signboard had been seen at least once, its information remained available for the rest of the episode, enabling the LLM to reason about stores that were not currently in view but had been observed earlier. This decoupling of perception (signboard parsing) and reasoning (LLM) made it easy to run multiple missions over the same world without re-engineering the mapping component.

\subsection{LLM Policy and Historical Log}
The LLM navigation policy receives two main inputs: the current semantic map and a history log describing which junctions and store instances have already been visited and in what order. Given a user request (for example, a shopping list over item categories), the navigation node constructs a structured prompt from these inputs and asks the LLM to output a discrete high-level action in the constrained format \texttt{<DIRECTION>|||<STORE\_ACTION>}.

We log every LLM query and response to a JSON file, which allows us to reconstruct the sequence of decisions taken during an episode. From these logs, we observe that early in the run the policy tends to choose actions that expand coverage of unexplored junctions, while later decisions focus on completing the remaining items on the shopping list and guiding the robot back toward the pickup area once all counts reach zero. Although no explicit optimality objective is enforced, the resulting sequences are generally reasonable and avoid re-entering already-completed store instances unless necessary.

\subsection{Grasping and Object Retrieval}
\begin{figure}[t]
    \centering
    {\includegraphics[width=\columnwidth]{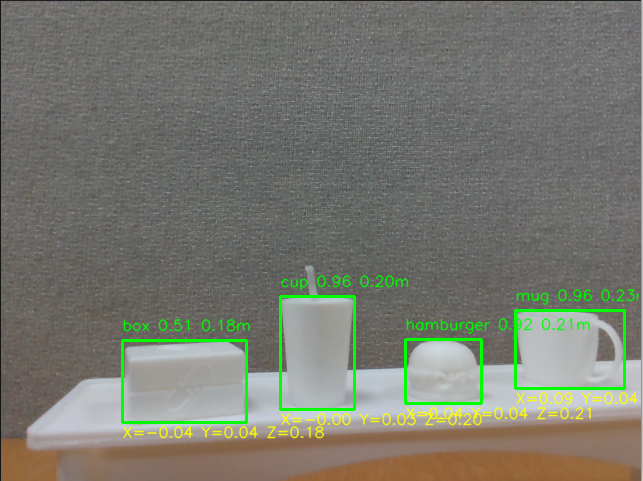}}
    \caption{\textbf{Example perception output for multi-object grasping.} The overlay reports class labels, detection confidence, and depth-based position estimates that guide the grasping controller.}
    \label{fig:screenshot_object_detection}
\end{figure}
\begin{figure}[t]
    \centering
    {\includegraphics[width=\columnwidth]{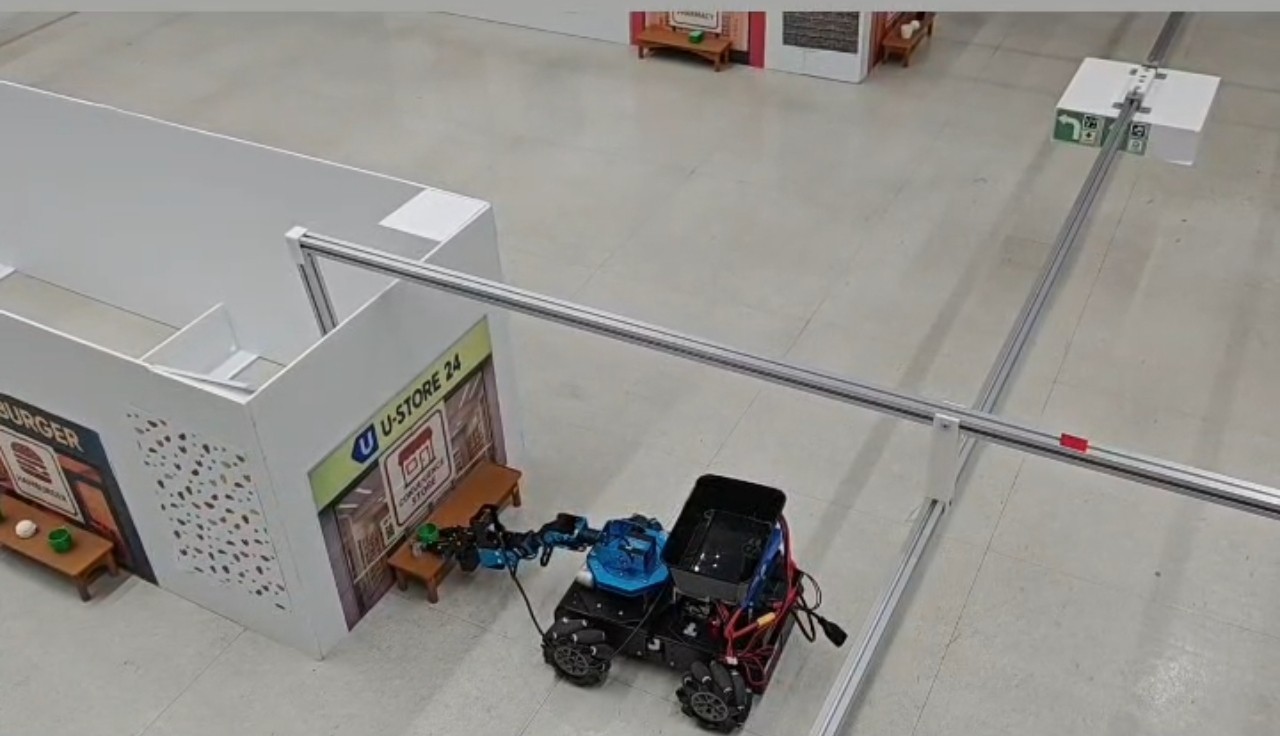}}
    \caption{\textbf{Snapshot of the robot performing object retrieval inside a store environment.} After base positioning via navigation primitives, the arm controller executes a predefined grasp sequence to pick up the requested item.}
    \label{fig:object_grasping}
\end{figure}

Grasping is triggered once the robot has been navigated to the vicinity of a target store or shelf. In our setup, the low-level navigation controllers are responsible for bringing the base to a suitable approach pose using AprilTag-based alignment and the local costmap, after which a dedicated grasping node (developed in the previous homework) takes over. As shown in Fig.~\ref{fig:screenshot_object_detection}, the grasping pipeline first produces class labels, confidence scores, and depth-based position estimates for the target objects, which are then used to align the end-effector before executing a predefined pickup motion.

In successful trials, the robot was able to approach the shelf, close the gripper around the target object, and back away without collision. Fig.~\ref{fig:object_grasping} shows a representative real-world demonstration of object retrieval after the robot enters a store. From the perspective of the LLM and navigation policy, grasping behaves as an atomic action: once a store is reached, the grasping node runs to completion and reports success or failure, and the high-level policy then decides which store or location to visit next.

\subsection{Failure Cases and Limitations}
Despite the overall success of the integrated system, several limitations were observed, mainly tag loss during approach and crowded or tightly spaced objects.

\subsubsection{Tag loss during approach}
AprilTag detections are occasionally lost due to viewing distance, angle, or partial occlusion by shelves and nearby objects. When this happens near the store entrance, the approach controller may not receive a stable relative pose, leading to aborted or inaccurate approach motions even though the local costmap still prevents hard collisions.

\subsubsection{Crowded or tightly spaced objects}
When target objects are placed too close to each other or to surrounding geometry, the predefined grasp motion may lead to fingertip collisions or unstable grasps. In some cases the robot can still pick up an object but with a marginal grip that is easily disturbed by subsequent motion or by slight pose errors from the base.

These failure modes suggest several directions for improvement, such as adding simple heuristics for re-acquiring tags, slowing down the base near shelves, or introducing a short local search to adjust the final base pose before triggering the grasping motion.

\section{Conclusion}
We presented an end-to-end ROS pipeline for LLM-driven agentic exploration in a corridor-based shopping task. The system integrates (i) perception and AprilTag-assisted localization, (ii) an online semantic map that records junction-to-store relations in a compact JSON format, (iii) an LLM decision layer that outputs constrained high-level actions, and (iv) a finite-state orchestrator that gates modular low-level skills for navigation, store entry, and grasping. Qualitative experiments in both Gazebo and a matching real-world layout demonstrate that the robot can execute full tasks from natural-language instructions through multi-store navigation to object retrieval.

Our modular separation between high-level semantic decisions and low-level deterministic controllers simplified development and improved execution stability, since policy changes did not require reworking motion control. Nevertheless, we observed practical failure modes such as intermittent tag loss near entrances and reduced grasp robustness when targets are crowded or tightly spaced. Future improvements include more reliable tag re-acquisition and entrance alignment, adaptive slowing and local search near shelves, and tighter coupling between uncertainty-aware perception and recovery behaviors.

\bibliographystyle{IEEEtran}
\bibliography{myblib}

@article{murartal2015orbslam,
  author  = {Mur-Artal, Ra{\'u}l and Montiel, J. M. M. and Tard{\'o}s, Juan D.},
  title   = {ORB-SLAM: A Versatile and Accurate Monocular SLAM System},
  journal = {IEEE Transactions on Robotics},
  year    = {2015},
  volume  = {31},
  number  = {5},
  pages   = {1147--1163},
  doi     = {10.1109/TRO.2015.2463671}
}

@article{murartal2017orbslam2,
  author  = {Mur-Artal, Ra{\'u}l and Tard{\'o}s, Juan D.},
  title   = {ORB-SLAM2: An Open-Source SLAM System for Monocular, Stereo, and RGB-D Cameras},
  journal = {IEEE Transactions on Robotics},
  year    = {2017},
  volume  = {33},
  number  = {5},
  pages   = {1255--1262},
  doi     = {10.1109/TRO.2017.2705103}
}

@article{campos2021orbslam3,
  author  = {Campos, Carlos and Elvira, Richard and G{\'o}mez Rodr{\'i}guez, Juan J. and Montiel, J. M. M. and Tard{\'o}s, Juan D.},
  title   = {ORB-SLAM3: An Accurate Open-Source Library for Visual, Visual--Inertial, and Multi-Map SLAM},
  journal = {IEEE Transactions on Robotics},
  year    = {2021},
  volume  = {37},
  number  = {6},
  pages   = {1874--1890},
  doi     = {10.1109/TRO.2021.3075644}
}

@inproceedings{olson2011apriltag,
  author    = {Olson, Edwin},
  title     = {AprilTag: A Robust and Flexible Visual Fiducial System},
  booktitle = {2011 IEEE International Conference on Robotics and Automation (ICRA)},
  year      = {2011},
  pages     = {3400--3407},
  doi       = {10.1109/ICRA.2011.5979561}
}

@inproceedings{wang2016apriltag2,
  author    = {Wang, John and Olson, Edwin},
  title     = {AprilTag 2: Efficient and Robust Fiducial Detection},
  booktitle = {2016 IEEE/RSJ International Conference on Intelligent Robots and Systems (IROS)},
  year      = {2016},
  doi       = {10.1109/IROS.2016.7759617}
}

@article{kostavelis2015semantic,
  author  = {Kostavelis, Ioannis and Gasteratos, Antonios},
  title   = {Semantic Mapping for Mobile Robotics Tasks: A Survey},
  journal = {Robotics and Autonomous Systems},
  year    = {2015},
  volume  = {66},
  pages   = {86--103},
  doi     = {10.1016/j.robot.2014.12.006}
}

@article{garg2021semantics,
  author       = {Garg, Sourav and S{\"u}nderhauf, Niko and Dayoub, Feras and Morrison, Douglas and Cosgun, Akansel and Carneiro, Gustavo and Wu, Qi and Chin, Tat{-}Jun and Reid, Ian D. and Gould, Stephen and Corke, Peter and Milford, Michael},
  title        = {Semantics for Robotic Mapping, Perception and Interaction: A Survey},
  journal      = {CoRR},
  volume       = {abs/2101.00443},
  year         = {2021},
  url          = {https://arxiv.org/abs/2101.00443},
  eprinttype   = {arXiv},
  eprint       = {2101.00443}
}

@inproceedings{huang2022zeroshot,
  author    = {Huang, Wenlong and Abbeel, Pieter and Pathak, Deepak and Mordatch, Igor},
  title     = {Language Models as Zero-Shot Planners: Extracting Actionable Knowledge for Embodied Agents},
  booktitle = {International Conference on Machine Learning (ICML)},
  series    = {Proceedings of Machine Learning Research},
  volume    = {162},
  year      = {2022},
  pages     = {9118--9147},
  publisher = {PMLR},
  url       = {https://proceedings.mlr.press/v162/huang22a.html}
}

@article{ahn2022saycan,
  author     = {Ahn, Michael and Brohan, Anthony and Brown, Noah and Chebotar, Yevgen and Cortes, Omar and David, Byron and Finn, Chelsea and Fu, Charlie and Gopalakrishnan, Keerthana and Hausman, Karol and Herzog, Alex and Ho, Daniel and Hsu, Jasmine and Ichter, Brian and Irpan, Alex and Jang, Eric and Jesmonth, Simon and Joshi, Nikhil and Julian, Ryan and Kalashnikov, Dmitry and Kuang, Yuhang and Le, Quoc V. and Lee, Linda and Levine, Sergey and Lu, Yao and Mourad, Nair and Omoindrot, R{\'e}mi and Rao, Kanishka and Reymann, Kasey and Riano, Lorenzo and Ryoo, Michael S. and Salakhutdinov, Ruslan and Sievers, Noah and Tan, Tolly and Tucker, George and Vanhoucke, Vincent and Xia, Fei and Zeng, Andy and Zhang, Peter and Zhu, Bichen},
  title      = {Do As I Can, Not As I Say: Grounding Language in Robotic Affordances},
  journal    = {arXiv preprint},
  year       = {2022},
  eprint     = {2204.01691},
  eprinttype = {arXiv},
  url        = {https://arxiv.org/abs/2204.01691}
}

@inproceedings{liang2023code,
  author    = {Liang, Jacky and Huang, Wenlong and Xia, Fei and Xu, Peng and Hausman, Karol and Ichter, Brian and Florence, Pete and Zeng, Andy},
  title     = {Code as Policies: Language Model Programs for Embodied Control},
  booktitle = {2023 IEEE International Conference on Robotics and Automation (ICRA)},
  year      = {2023},
  pages     = {9493--9500},
  doi       = {10.1109/ICRA48891.2023.10160591}
}

@inproceedings{quigley2009ros,
  title        = {{ROS}: an open-source Robot Operating System},
  author       = {Quigley, Morgan and Conley, Ken and Gerkey, Brian and Faust, Josh and Foote, Tully and Leibs, Jeremy and Wheeler, Rob and Ng, Andrew Y.},
  booktitle    = {ICRA Workshop on Open Source Software},
  year         = {2009}
}

@inproceedings{koenig2004gazebo,
  title        = {Design and Use Paradigms for {Gazebo}, an Open-Source Multi-Robot Simulator},
  author       = {Koenig, Nathan and Howard, Andrew},
  booktitle    = {IEEE/RSJ International Conference on Intelligent Robots and Systems (IROS)},
  year         = {2004}
}

@misc{bai2025qwen25vltechnicalreport,
      title={Qwen2.5-VL Technical Report}, 
      author={Shuai Bai and Keqin Chen and Xuejing Liu and Jialin Wang and Wenbin Ge and Sibo Song and Kai Dang and Peng Wang and Shijie Wang and Jun Tang and Humen Zhong and Yuanzhi Zhu and Mingkun Yang and Zhaohai Li and Jianqiang Wan and Pengfei Wang and Wei Ding and Zheren Fu and Yiheng Xu and Jiabo Ye and Xi Zhang and Tianbao Xie and Zesen Cheng and Hang Zhang and Zhibo Yang and Haiyang Xu and Junyang Lin},
      year={2025},
      eprint={2502.13923},
      archivePrefix={arXiv},
      primaryClass={cs.CV},
      url={https://arxiv.org/abs/2502.13923} 
}

@misc{yolo11_ultralytics,
  author = {Glenn Jocher and Jing Qiu},
  title = {Ultralytics YOLO11},
  version = {11.0.0},
  year = {2024},
  url = {https://github.com/ultralytics/ultralytics},
  orcid = {0000-0001-5950-6979, 0000-0003-3783-7069},
  license = {AGPL-3.0}
}
\end{document}